\documentclass[conference]{IEEEtran}
\makeatletter\if@twocolumn\PassOptionsToPackage{switch}{lineno}\else\fi\makeatother
\usepackage{bigdelim}

\usepackage{graphicx}
\usepackage[T1]{fontenc}
\usepackage[numbers,sort&compress]{natbib}
\usepackage{mathtools}
\usepackage{booktabs}
\usepackage{lineno,hyperref}
\usepackage{amsmath}
\usepackage{amsfonts}
\usepackage{amssymb}
\usepackage{amsmath}
\usepackage{multirow,morefloats,floatflt,cancel,tfrupee}
\usepackage{makecell}
\usepackage{tikz}
\usepackage{setspace}
\usepackage{color}
\usepackage{multirow}
\usepackage{adjustbox}
\usepackage{bbding}
\usepackage{physics}
\usepackage{colortbl}
\usepackage{xcolor}
\usepackage{pifont}
\usepackage[nointegrals]{wasysym}
\renewcommand\IEEEkeywordsname{Keywords}

\makeatletter

\let\old@ps@IEEEtitlepagestyle\ps@IEEEtitlepagestyle
\def\confheader#1{%
    \def\ps@IEEEtitlepagestyle{%
        \old@ps@IEEEtitlepagestyle%
        \def\@oddhead{\strut\hfill#1\hfill\strut}%
        \def\@evenhead{\strut\hfill#1\hfill\strut}%
    }%
    \ps@headings%
}
\makeatother

\begin{document}

        \title{Improved EATFormer: A Vision Transformer for Medical Image Classification}
\author{\IEEEauthorblockN{Yulong Shisu,
Susano Mingwin,
Yongshuai Wanwag,
Zengqiang Chenso,
Sunshin Huing}\\

University of Chinese Academy of Sciences. Northwestern University}

\maketitle

\begin{abstract}

The accurate analysis of medical images is vital for diagnosing and predicting medical conditions. Traditional approaches relying on radiologists and clinicians suffer from inconsistencies and missed diagnoses. Computer-aided diagnosis systems can assist in achieving early, accurate, and efficient diagnoses. This paper presents an improved Evolutionary Algorithm-based Transformer architecture for medical image classification using Vision Transformers.
The proposed EATFormer architecture combines the strengths of Convolutional Neural Networks and Vision Transformers, leveraging their ability to identify patterns in data and adapt to specific characteristics. The architecture incorporates novel components, including the Enhanced EA-based Transformer block with Feed-Forward Network, Global and Local Interaction , and Multi-Scale Region Aggregation modules. It also introduces the Modulated Deformable MSA module for dynamic modeling of irregular locations.
The paper discusses the Vision Transformer (ViT) model's key features, such as patch-based processing, positional context incorporation, and Multi-Head Attention mechanism. It introduces the Multi-Scale Region Aggregation module, which aggregates information from different receptive fields to provide an inductive bias. The Global and Local Interaction module enhances the MSA-based global module by introducing a local path for extracting discriminative local information.
Experimental results on the Chest X-ray and Kvasir datasets demonstrate that the proposed EATFormer significantly improves prediction speed and accuracy compared to baseline models.
\end{abstract}

\begin{IEEEkeywords}
Vision Transformer, Medical Image Classification, Deep Learning, Medical Imaging.
\end{IEEEkeywords}

\section{Introduction}

The precise examination of medical images plays a crucial role in the diagnosis and prediction of various medical conditions. Timely detection of lesions, tumors, and other abnormalities is vital in identifying life-threatening diseases and enabling prompt and effective treatment. In the past, medical image interpretations were primarily performed by radiologists and clinicians. However, this approach is prone to inconsistencies between observers, fatigue, and a significant number of missed diagnoses~\cite{wallace2022impact}. Consequently, clinicians cannot consistently achieve high levels of accuracy when analyzing and interpreting medical data. Moreover, chest X-rays can be employed to diagnose a range of disorders, including lung infections, COVID-19 (SARS-CoV-2), pneumonia, heart disease, and injuries. Obtaining early and accurate diagnoses for these lung conditions is crucial for effective treatment. Gastrointestinal (GI) disorders encompass diseases that affect the digestive tract, such as the esophagus, stomach, small intestine, and colon. These ailments can vary from benign conditions like acid reflux to potentially fatal conditions like colon cancer. Early detection and accurate diagnosis are essential for timely and effective treatment, leading to improved patient outcomes and a higher quality of life. Therefore, the implementation of a computer-aided diagnosis (CAD) system could assist physicians and clinical experts in achieving early, accurate, and efficient diagnoses.

Convolutional Neural Networks (CNNs) possess an inherent inclination for identifying patterns in data, and their convolution operations can be adjusted to suit the specific characteristics of the data. This amalgamation of inductive bias and adaptability empowers CNNs to excel in tasks associated with computer vision. Similarly, Vision Transformers have emerged as innovative solutions for visual tasks within the realm of computer vision. These networks now find numerous real-world applications, such as autonomous driving~\cite{zhang2022cctsdb}, object detection~\cite{pyramid}, health care~\cite{manzari2024befunet}, and defense-related applications. The current study specifically concentrates on methods for medical image classification. More precisely, four variants of Vision Transformers are employed: the Data-Efficient Image Transformer (DeiT)~\cite{Deit}, robust vision transformer (MedViT)~\cite{medvit}, localvit~\cite{li2021localvit}, and Swin Vision Transformer~\cite{swin}. In addition, six convolutional neural networks are used as baseline models: Xception~\cite{xception}, ResNet~\cite{Resnet}, DenseNet~\cite{VGG16}, MobileNet~\cite{Mobilenets}, EfficientNet~\cite{efficientnet}, and Inception~\cite{inception}.

This paper presents a rationale for the applicability of Vision Transformer by drawing an analogy with the well-established Evolutionary Algorithm (EA), highlighting their shared mathematical formulation. Building on the effectiveness of various EA variants, we introduce a novel pyramid EATFormer backbone, which exclusively incorporates the proposed EA-based Transformer (EAT) block. The EAT block comprises three enhanced parts: Feed-Forward Network (FFN), Global and Local Interaction (GLI), and Multi-Scale Region Aggregation (MSRA) modules. These components are designed to independently capture interactive, individual, and multi-scale, information. Additionally, As part of the transformer backbone, we develop a Task-Related Head (TRH) that enables flexible information fusion at the end of the process. Furthermore, we enhance the approach with a Modulated Deformable MSA (MD-MSA) module for dynamic modeling of irregular locations. As demonstrated by our experimental results on the Chest X-ray dataset~\cite{cohen2020covid} and Kvasir~\cite{pogorelov2017kvasir} dataset, our approach significantly improves both prediction speed and accuracy.

 \begin{figure*}[!t]
 \centering
  \includegraphics[width=\textwidth]{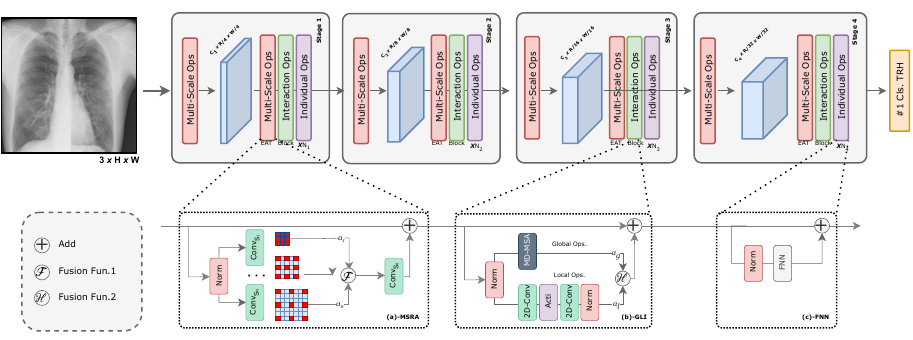}
  \caption{Network architecture of the proposed Transformer.}\label{fig.architecure}
\end{figure*}

\section{The Proposed Approach}

The improved EATFormer architecture, depicted in Figure 1, consists of four stages with varying resolutions, following the hierarchical structure~\cite{PVT}. It incorporates EAT blocks, which include three mixed paradigm $y=f(x)+x$ residuals: (a) Feed-Forward Network (FFN), (b) Global and Local Interaction (GLI), and (c) Multi-Scale Region Aggregation (MSRA) modules. The down-sampling procedure between two stages is achieved using MSRA with a stride greater than 1. Additionally, we introduce a novel Modulated Deformable MSA (MDMSA) for enhanced global modeling and a Task-Related Head (TRH) that provides a more elegant and flexible approach to the classification task.

\subsection{Overview of Vision Transformer}

The ViT model~\cite{Vit} utilizes a step-by-step process to effectively process input images. Initially, it employs a technique where each input image is divided into multiple non-overlapping tokens, also known as patches, with a predetermined fixed length. These patches are subsequently subjected to a trainable linear projection layer, which sequentially expands and contracts the embedding dimension. This process aims to extract meaningful information from each patch and ensure compatibility with subsequent operations.
ViT's architecture incorporates the positional context of every patch as one of its distinctive features. Embeddings of position information are attached to deformed patches in order to accomplish this. Position embeddings are vital for providing contextual information pertaining to patch relative and absolute positions in an image. By incorporating positional context, the ViT model enhances its understanding of the spatial relationships within the input data.
Furthermore, the ViT model integrates an additional classification token called CLS into the embedded patches. This token serves as a representation of the entire image and is vital for downstream tasks, particularly image classification.

The tokens, including both patches and the CLS token, are then fed into a encoder for transforming process. The Transformer encoder consists of multi-layer perceptron (MLP) layers and the Multi-Head Attention (MSA). Through the Multi-Head Attention mechanism, the embedded patches are divided into multiple heads, enabling each head to learn distinct states of self-attention. This facilitates capturing various aspects and relationships within the input data. The outputs from all the heads are subsequently merged and forwarded to a MLP layer for further processing.
To ensure stable information flow, normalization layers and residual blocks are applied after and before each MLP and MSA operation, respectively. These layers contribute to the overall stability and optimization of the model.
the ViT model adopts a comprehensive approach by leveraging patch-based processing, positional context incorporation, and a combination of MSA and MLP operations within the Transformer encoder. This approach allows the model to effectively process images, enabling tasks such as image classification to be performed accurately. In the Transformer encoder, a sequence of inputs $u$ is entered and the outputs $v$ are calculated as follows:

\begin{equation}\label{eq.1}
 \hat{u} = u + \operatorname{MSA}(\operatorname{Norm}(u))
\end{equation}

\begin{equation}\label{eq.2}
 v = \hat{u} + \operatorname{MLP}(\operatorname{Norm}((\hat{u}))
\end{equation}

\noindent where $\operatorname{Norm}$ is the normalization of layers, $\operatorname{MLP}$ is the multilayer perceptron, and $\operatorname{MSA}$ is the multi-head self-attention. After the transformer encoder, ViT utilizes the CLS for the final prediction phase.

\subsection{Multi-Scale Region Aggregation}
As inspired by some evolutionary algorithm (EA) methods~\cite{li2021differential, zhang2022eatformer} that utilize multiple populations and different searching regions in order to improve model performance, we apply the same concept to 2D image analysis. As part of our study of a vision transformer, we introduce a novel module called Multi-Scale Region Aggregation (MSRA). In Figure 4.(a), we illustrate the structure of MSRA, which incorporates $N$ local convolution operations (Conv $_{S_n}, 1 \leq n \leq N$) with varying strides. By aggregating information from different receptive fields, these operations effectively provide an inductive bias without requiring additional position embedding procedures. Specifically, the $n$-th dilation operation $o_n$, which transforms the input feature map $x$, can be mathematically expressed as follows:

$$
\begin{aligned}
& o_n(\boldsymbol{x})=\operatorname{Conv}_{S_n}(\operatorname{Norm}(\boldsymbol{x})) \\
& \text { s.t. } n=1,2, \ldots, N,
\end{aligned}
$$

To mix all operations in a weighted manner, we introduce the Weighted Operation Mixing (WOM) mechanism, which uses a softmax function over a set of learnable weights $\alpha_1, \ldots, \alpha_N$. As a result of applying the mixing function $\mathcal{F}$, the intermediate representation $\boldsymbol{x}_o$ can be obtained as follows:

$$
x_o=\sum_{n=1}^N \frac{\exp \left(\alpha_n\right)}{\sum_{n^{\prime}=1}^N \exp \left(\alpha_{n^{\prime}}\right)} o_n(\boldsymbol{x}) \text {, }
$$

In the given formula, $F$ represents the addition function. There are other fusion functions available, such as concatenation, which yields better results, but come with additional parameters. For the sake of simplicity, the paper opts to use the addition function by default. Subsequently, a convolution layer, ($Conv_{So}$), is employed to map the intermediate representation, $x_o$, to the same number of channels as the input, $x$. Relative connections are used to obtain the module's final output. The MSRA module also contributes to the uniformity and elegance of the EATFormer, as it serves as the embedding patch and stem. Because of its CNN-based MSRA, this paper does not use any form of position embedding since the GLI module inherently provides a natural inductive bias.

\subsection{Global and Local Interaction}

Influenced by the introduction of local search procedures in EA variants to enhance the convergence of high-quality solutions \cite{li2021localvit, medvit} in a faster and more effective manner (refer to Figure 1-(c) for a clearer visual explanation), we propose an enhancement to the MSA-based global module, transforming it into a novel module called Global and Local Interaction (GLI) module. Figure 4-(b) illustrates the GLI module, which incorporates an additional local path alongside the global path. The local path aims to extract more discriminative information relevant to localities, similar to the concept of a local population mentioned earlier, while the global path retains its focus on modeling global information. At the channel level, input features are divided into local features (indicated in blue) and global features (indicated in green). These features are then separately processed by the global and local paths to facilitate feature interactions. To ensure a balanced contribution from both paths, we employ the proposed Weighted Operation Mixing mechanism described in Section 4.3.1, which adjusts the weights assigned to the local branch ($\alpha_l$) and the global branch ($\alpha_g$). The outputs from these two paths are then combined, resulting in data with the original dimensions. This combination can be viewed as a flexible plug-and-play concatenation operation denoted as $H$, thus serving as an improved module for the current transformer structure. It's worth noting that the local operation can take various forms, such as a traditional convolution layer or other enhanced modules like DCN~\cite{zhu2019deformable} or local MSA. On the other hand, the global operation can utilize MSA, D-MSA, Performer~\cite{choromanski2020rethinking}, and other similar techniques.

In this research paper, we have selected naive convolution with MSA modules as the fundamental building blocks of the GLI module. GLI's choice ensures that global modeling capability is maintained while locality is enhanced, as illustrated in Figure 1. Notably, the maximum path length between any two positions in our proposed structure remains $O(1)$, which helps preserve efficiency and parallelism, similar to the vanilla ViT.
The selection of the feature separation ratio, denoted as $p$, is a crucial factor influencing the efficiency and effectiveness of the model. Different ratios result in varying parameters, FLOPs (floating-point operations), and precision of the model. To provide a detailed explanation, the local path consists of a set of point-wise feature maps in the dimensions of $R^{C\times H\times W} = R^{C\times L}$, while both paths involve $k \times k$ depth-wise convolutions. Assuming that the number of global channels is $C_g = p \times C$ and the number of local channels is $C_l = C - C_g$.
Now, let us present an analysis of the parameter count and computational aspects of the improved GLI module, which are as follows:
$$
\begin{aligned}
\text { Params }= & 5 C_g^2+\left(2-2 C-K^2\right) C_g+ \\
& \left(k^2+2+C\right) C .
\end{aligned}
$$

 \begin{figure}[!t]
 \centering
  \includegraphics[width=\linewidth]{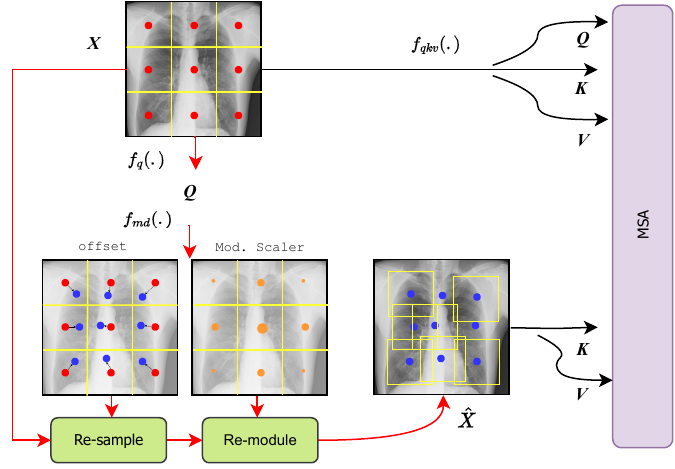}
  \caption{Structure of the proposed MD-MSA.}\label{fig.module}
\end{figure}

\subsection{Modulated Deformable MSA}

Taking inspiration from the non-horizontal and non-vertical spatial distribution observed in real individuals, we have made enhancements to a novel module called Modulated Deformable MSA (MD-MSA). This module takes into account the fine-tuning of positions and the re-weighting of each spatial patch. Figure 2 illustrates the comparison between the naive MSA procedure (represented by the blue dotted line) and the MD-MSA procedure (represented by the red solid line).
In the naive MSA procedure, the QKV features are obtained from the input feature map $X$ using the function fqkv(.), expressed as $Q K V=f_{q k v}(\boldsymbol{X})$, where $f_{q k v}=f_q \oplus f_k \oplus f_v(\oplus$ is a concatenation operation combining the functions $f_q$, $f_k$, and $f_v$. On the other hand, the MD-MSA procedure introduces modifications. The key distinction between the proposed MD-MSA and the original MSA lies in the query-aware access of the fine-tuned feature map $\hat{X}$ to extract $\boldsymbol{K}$ and $\boldsymbol{V}$ features.
To elaborate further, when given an input feature map $X$ with $L$ positions, the function $f_q$ is applied to obtain the query matrix $Q$, denoted as $\boldsymbol{Q}=f_q(\boldsymbol{X})$. This query matrix is then used to predict deformable offsets $\Delta l$ and modulation scalars $\Delta m$ for all positions in order to access the fine-tuned feature map $\Delta m$:
$$
\Delta l, \Delta m=f_{m d}(\boldsymbol{Q})
$$

Using the $l$-th position, we calculate the resampled and reweighted feature $\hat X_l$ as follows:

$$
\hat{\boldsymbol{X}}_l=\mathcal{S}\left(\boldsymbol{X}_l, \Delta l\right) \cdot \Delta m
$$

where $\Delta l$ represents the relative coordinate with an unconstrained range for the $l$-th position, while $\Delta m$ is confined within the range $(0,1)$. The symbol $\mathcal{S}$ denotes the bilinear interpolation function. Consequently, $\boldsymbol{K} \boldsymbol{V}$ is derived from the new feature map $\operatorname{map} \hat{\boldsymbol{X}}$, denoted as $\boldsymbol{K} \boldsymbol{V}=f_{k v}(\hat{\boldsymbol{X}})$. MD-MSA differs from recently published work~\cite{xia2022vision} primarily in the modulation operation. The MD-MSA approach optimizes results by paying attention to a variety of positional characteristics.

\section{Experiments}

In this section, we initiate the discussion by examining the datasets utilized to train our enhanced Transformer model, and provide a summary of the experimental configurations. Following that, we conduct a comprehensive analysis of the experimental outcomes, wherein we compare the performance of our Transformer model with that of cutting-edge studies in the domain of medical image classification.

\subsection{Datasets and Implementation details}

To assess the performance of the model architecture, we utilized two publicly available medical datasets representing various modalities. The first dataset we employed is the \textbf{Chest X-ray dataset} \cite{cohen2020covid}, comprising 7135 chest X-ray images from patients. Among these images, 576 depict COVID-19 cases, 1583 are classified as normal, 4273 show pneumonia, and 703 exhibit tuberculosis. The images were collected from different publicly accessible resources. Our second dataset, called \textbf{Kvasir}~\cite{pogorelov2017kvasir}, is a multi-class dataset containing 1000 images per class, resulting in a total of 8000 images distributed across eight distinct classes. These classes encompass pathological findings (esophagitis, polyps, ulcerative colitis), anatomical landmarks (z-line, pylorus, cecum), as well as normal and regular findings (normal colon mucosa, stool), and polyp removal cases (dyed and lifted polyps, dyed resection margins).

To thoroughly evaluate the effectiveness and dependability of our model for medical image classification, we conducted training and evaluation using the Kvasir and Chest X-ray datasets. Throughout the training process, we utilized the Adam optimizer with a learning rate of 0.005 and a batch size of 100. Various data augmentation techniques, including random rotations, zooming, and horizontal flips, were incorporated to augment the training set and bolster the model's ability to generalize. The experiments were conducted on a Jupyter-notebook equipped with a GeForce RTX 3080 Ti GPU, and the primary code implementation was carried out using Python 3.7.

\subsection{Evaluation Measures}

To assess the effectiveness of classification models, key performance indicators (KPIs) commonly used are Matthews correlation coefficient (MCC), f1-score, accuracy, recall, and precision. The accuracy metric evaluates the overall correctness of the model's predictions. It is determined by dividing the number of correctly classified samples by the total number of samples, as shown in Equation (3). Precision measures the proportion of true positives among all positive predictions made by the model, as indicated in Equation (4). Recall calculates the proportion of true positives among all actual positive samples in the dataset, as illustrated in Equation (5). The f1-score, which is the harmonic mean of precision and recall, provides a balanced assessment combining these two metrics, as demonstrated in Equation (6). In our experimentation, we will utilize these key performance indicators (KPIs) to assess the effectiveness of our proposed model. By employing these metrics, we can evaluate the model's capability to accurately classify medical images.

\begin{equation}
A C C =\frac{T_P+T_N}{T_P+T_N+F_P+F_N}\\
\end{equation}
\begin{equation}
\text { recall } =\frac{T_P}{T_P+F_N} \\
\end{equation}
\begin{equation}
\text { precision } =\frac{T_P}{T_P+F_P} \\
\end{equation}
\begin{equation}
F_{1}-\text { score } =\frac{2 \times \text { precision } \times \text { recall }}{\text { precision }+\text { recall }}
\end{equation}
\vspace{2mm}

\begin{table}[!t]
\centering
\caption{Quantitative results on Chest X-ray\cite{cohen2020covid} dataset.}
\begin{adjustbox}{width=.95\linewidth,center}
    \begin{tabular}{l|l|l|l|l|l|l}
    \toprule
     \textbf{Method} & \textbf{Precision} &\textbf{Recall} & \textbf{F1-score} & \textbf{Accuracy}   & \textbf{MCC}& \textbf{FPS}\\ \midrule

           DenseNet   & 0.9345  & 0.9326  & 0.9319  & 0.9326 &  0.8931 &  16.08\\\hline

           ResNetV2  &  0.9169  &  0.9157   & 0.9150   &0.9157 & 0.8658 &  22.47 \\ \hline

           Xception  & 0.9380  &  0.9377  &  0.9375  &  0.9377 & 0.9010 &  21.34 \\  \hline

           EfficientNet   &  0.9314  &  0.9287   & 0.9277   & 0.9287 & 0.8878 & 26.15\\ \hline

           Inception & 0.9415   &  0.9403  & 0.9400  & 0.9287 &  0.9052  &  20.48\\ \hline

           MobileNetV2 & 0.9133 &  0.9092  &  0.9099  &  0.9092 & 0.8567 &  21.85\\ \hline

           DeiT-Ti  &   0.9266  &  0.9261  &  0.9262  &  0.9261 & 0.8821 &  25.48 \\ \hline

           Localvit  &  0.9332 &  0.9300  &  0.9291&  0.9300 & 0.8899 &  16.75 \\ \hline

           Swin  & 0.9291  & 0.9248   & 0.9248  & 0.9248 & 0.8813  &  20.74 \\  \hline

           MedViT &   \textcolor{blue}{0.9527} & \textcolor{blue}{0.9520} & \textcolor{blue} {0.9521} & \textcolor{blue}{0.9520} &  \textcolor{blue}{0.9236} &  {11.49} \\ \hline

           Ours &  \textcolor{red}{0.9533}   & \textcolor{red}{0.9533} &  \textcolor{red}{0.9532} & \textcolor{red}{0.9533} &  \textcolor{red}{0.9259} &  {11.20}\\
            \hline

    \end{tabular}
    \end{adjustbox}
    \label{table: Result chest xra}
\end{table}

\begin{table}[!t]
\caption{Quantitative results on the Kvasir~\cite{pogorelov2017kvasir} dataset.}
\begin{adjustbox}{width=.95\linewidth,center}
 \label{table:KvasirCapsule}
     \begin{tabular}{c|c|c|c|c|c|c|c}
     \toprule
     \textbf{Method} & \textbf{Precision} &\textbf{Recall} & \textbf{F1-score} & \textbf{Accuracy}   & \textbf{MCC}& \textbf{FPS}\\ \midrule

          DenseNet  & 0.9284  & 0.9263  &0.9258 & 0.9263 &  0.9161 &  12.32\\\hline

          ResNetV2  & 0.9262 &  0.9250 & 0.9246  & 0.9250  &  0.9146 &  21.59 \\ \hline

          Xception &  0.9032  & 0.9025  & 0.9020  & 0.9025  & 0.8888 &  43.12  \\  \hline

          EfficientNet  &  0.9136  &  0.9112   & 0.9105  &0.9113 & 0.8991 &  24.35 \\ \hline

          Inception & 0.8877    &   0.8875   & 0.8870  & 0.8875 &  0.8716  & 21.18 \\ \hline

          MobileNetV2  & 0.8789   &  0.8775 &  0.8769  &  0.8775 & 0.8603 & 43.97\\ \hline

          DeiT-Ti  & 0.9353  & 0.9350 & 0.9349   & 0.9350 &0.9258 &  25.80 \\  \hline

          Localvit  &0.9375   & 0.9337  & 0.9333  & 0.9337 & 0.9249 &  13.35 \\\hline

          Swin  & \textcolor{blue}{0.9433}   & 0.9400   & 0.9398  & 0.9400 & 0.9316 &  21.50 \\ \hline

          MedViT & 0.9408  & \textcolor{blue} {0.9401}  & \textcolor{blue}{0.9399}  & \textcolor{blue}{0.9401}  & \textcolor{blue}{0.9319} &  15.81\\  \hline

          Ours & \textcolor{red}{0.9454 }  & \textcolor{red}{0.9437} &  \textcolor{red}{0.9436}  & \textcolor{red}{0.9437} &  \textcolor{red}{0.9360} & 11.34\\  \hline

     \end{tabular}
     \end{adjustbox}
\label{table:Kvasir result}
\end{table}

\subsection{Comparison with the state-of-the-art}

\textbf{Chest X-ray dataset:} Table 1 displays the evaluation outcomes of the proposed model and several advanced classification algorithms applied to the Chest X-ray dataset.
The table reveals that our Transformer model exhibits significant advantages in medical image classification. Similar to Swin and NextViT Transformer, we establish a hierarchical structure to enhance the feature representation capabilities of neural networks across various scales. However, our network achieves 2.85\% and 0.13\% improvement in classification accuracy compared to NextViT and Swin Transformer, with respective accuracy rates of 92.48\% and 95.20\%, as well as a 2.33\% improvement compared to LNL, which attains an accuracy of 93.00\% (compared to 95.33\%). While LNL adopts a multi-branch structure that combines the strengths of CNN and Transformer, our approach differs by fusing the branches of the EAT-block at different levels to reduce computational complexity while enhancing the accuracy of medical image classification.
These experimental findings further validate that hierarchically fusing feature information from distinct branches can lead to varying degrees of reduced computational complexity and improved performance of the classification model. Our model embodies both of these characteristics.

\textbf{Kvasir:} In order to assess the effectiveness and robustness of the proposed model for medical image classification, The proposed model was compared with several well-known models, namely Swin, LNL, DenseNet, and EfficientNet on the Kvasir dataset.
Based on Table 2, the proposed model demonstrates better classification performance than the other methods. Based on the results, it is evident that the proposed model outperforms all other models on the Kvasir dataset while utilizing the smallest number of parameters. According to the study, the system scored higher on F1-score, accuracy, recall, precision compared with Swin, NextViT, LNL, DenseNet, and EfficientNet. Specifically, the proposed model achieved an accuracy rate of 94.37\%, surpassing DenseNet by 1.74 percentage points, Xception by 4.12 percentage points, EfficientNet by 3.24 percentage points, and Inception by 5.62 percentage points. Moreover, the proposed model achieved a precision of 94.54\%, outperforming NextViT and Swin, which achieved precisions of 94.33\% and 94.08\%, respectively. Furthermore, in terms of recall, the proposed model exhibited superior performance with a score of 94.37\%, surpassing NextViT (94.00\%) and Swin (94.01\%).

According to the results presented in Tables I and II, it can be observed that our transformer architecture surpassed other pre-trained CNN and Transformers models across all metrics. This finding suggests that the Multi-Scale Region Aggregation employed by our transformer model enables it to effectively capture multi-scale patterns in the data, leading to more accurate predictions of the images. Moreover, the sequential processing of data in visual transformers allows for efficient parameter utilization, consideration of longer-term dependencies, and the acquisition of more generalized knowledge across diverse inputs. Additionally, the absence of recurrence in transformers contributes to improved generalization and diminishes the likelihood of overfitting.

\section{Conclusion}

This paper presented a novel approach for medical image classification using the Vision Transformer architecture. The proposed EATFormer architecture incorporated evolutionary algorithm-inspired modules such as Multi-Scale Region Aggregation (MSRA), Global and Local Interaction (GLI), and Modulated Deformable MSA (MD-MSA) to enhance the performance of the model. The experimental results on the Chest X-ray dataset and Kvasir dataset demonstrated significant improvements in both prediction speed and accuracy compared to baseline models. The study showcased the effectiveness of Vision Transformers in medical image analysis and highlighted the potential of incorporating evolutionary algorithm concepts into deep learning architectures. The proposed approach has the potential to assist physicians and clinicians in achieving early, accurate, and efficient diagnoses, leading to improved patient outcomes and a higher quality of life.

\bibliographystyle{IEEEtran}

\bibliography{article}

\end{document}